\documentclass[twoside,leqno,twocolumn]{article}

\usepackage[letterpaper]{geometry}
\usepackage{paralist}
\usepackage{ltexpprt}
\usepackage{hyperref}

\usepackage{algorithm, algpseudocode}
\usepackage{algorithmicx}
\usepackage{amsmath,amssymb,amsfonts}
\usepackage{xspace}
\usepackage[inline]{enumitem}
\usepackage{graphicx}
\usepackage{tabularx}
\usepackage[multiple]{footmisc}
\usepackage{subcaption}
\newcommand{\fgr}[3][\relax]{%
	\begin{figure}[htp]%
		\centering
		\includegraphics[#2]{#3}%
		\ifx\relax#1\else\caption{{#1}}\fi
	\end{figure}%
}

\usepackage{xcolor}

\newcommand{\method}{{\sc C-AllOut}\xspace}

\definecolor{OliveGreen}{rgb}{0,0.6,0}

\newcommand{\cbit}{\begin{compactitem}}
	\newcommand{\ceit}{\end{compactitem}}
\newcommand{\cben}{\begin{compactenum}}
	\newcommand{\ceen}{\end{compactenum}}

\newcommand{\bal}{\begin{align}}
\newcommand{\ean}{\end{align}}
\newcommand{\bit}{\begin{itemize}}
\newcommand{\eit}{\end{itemize}}
\newcommand{\ben}{\begin{enumerate}}
\newcommand{\een}{\end{enumerate}}
\newcommand{\beq}{\begin{equation}}
\newcommand{\eeq}{\end{equation}}

\newcommand{\bE}{\mathbf{E}}
\newcommand{\bN}{\mathbf{N}}

\makeatletter
\algnewcommand{\LineComment}[1]{\Statex \hskip\ALG@thistlm \(\triangleright\) #1}
\makeatother

\newcommand{\reminder}[1]{{\textsf{\textcolor{red}{[#1]}}}}
\newcommand{\la}[1]{{\textsf{\textcolor{purple}{[LA: #1]}}}}
\newcommand{\rc}[1]{{\textsf{\textcolor{orange}{[RC: #1]}}}}

\newcommand{\hide}[1]{}

\newtheorem{ranking}{Ranking}

\newcolumntype{?}{!{\vrule width 2pt}}

\newcommand{\dataset}{\mathbf{X}}
\newcommand{\point}[1]{\mathbf{x}_{#1}}
\newcommand{\pointref}[2]{\mathbf{x}^{#1}_{#2}}
\newcommand{\pointi}{\point{i}}
\newcommand{\pointj}{\point{j}}
\newcommand{\dist}[2]{\mathtt{d}(#1, #2)}
\newcommand{\distij}{\dist{\pointi}{\pointj}}

\newcommand{\tree}{\mathcal{T}}
\newcommand{\node}{\eta}
\newcommand{\rootnode}{\rho}
\newcommand{\rootpoints}{\mathbf{R}}
\newcommand{\leafname}{\tau}
\newcommand{\leaf}[1]{\leafname^{(#1)}}
\newcommand{\leafi}{\leaf{i}}

\newcommand{\leafpointsname}{\mathbf{L}}
\newcommand{\leafpoints}[1]{\leafpointsname^{(#1)}}
\newcommand{\leafanypoints}{\leafpoints{\leafname}}
\newcommand{\leafipoints}{\leafpoints{i}}
\newcommand{\ant}[1]{\alpha_{#1}}
\newcommand{\antany}{\ant{\node}}
\newcommand{\anti}{\ant{\leafi}}

\newcommand{\rep}[1]{\pointref{(#1)}{\mathtt{e}}}
\newcommand{\repany}{\rep{\node}}
\newcommand{\repleaf}{\rep{\leafname}}
\newcommand{\repi}{\rep{\leafi}}
\newcommand{\croot}[1]{\pointref{(#1)}{\mathtt{r}}}
\newcommand{\crooti}{\croot{i}}
\newcommand{\crootj}{\croot{j}}
\newcommand{\frep}[1]{\pointref{(#1)}{\mathtt{f}}}
\newcommand{\frepi}{\frep{i}}
\newcommand{\frepj}{\frep{j}}
\newcommand{\nn}[1]{\pointref{(#1)}{\mathtt{nn}}}
\newcommand{\nnleaf}{\nn{\leafname}}
\newcommand{\nni}{\nn{i}}
\newcommand{\nnj}{\nn{j}}
\newcommand{\fn}[1]{\pointref{(#1)}{\mathtt{a}}}

\newcommand{\fnleaf}{\fn{\leafname}}

\newcommand{\distnnname}{\mathtt{d_{nn}}}
\newcommand{\distnn}[1]{\mathtt{d}^{(#1)}_{\mathtt{nn}}}
\newcommand{\distnni}{\distnn{i}}
\newcommand{\distnnj}{\distnn{j}}
\newcommand{\lradiusname}{\mathtt{a}}
\newcommand{\lradius}[1]{\lradiusname^{(#1)}}
\newcommand{\lradiusany}{\lradius{\leafname}}
\newcommand{\lradiusi}{\lradius{\leafi}}

\newcommand{\llradius}[1]{\lradiusname_{#1}}
\newcommand{\llradiusi}[1]{\lradiusname_{i}}
\newcommand{\llradiusj}[1]{\lradiusname_{j}}
\newcommand{\slradiuses}{\mathtt{A}}

\newcommand{\rank}{\mathcal{R}}
\newcommand{\orank}{\rank_o}
\newcommand{\grank}{\rank_g}
\newcommand{\lrank}{\rank_l}
\newcommand{\crank}{\rank_c}
\newcommand{\oscorename}{\mathtt{s_o}}
\newcommand{\oscore}[1]{\mathtt{s}^{(#1)}_{\mathtt{o}}}
\newcommand{\oscorei}{\oscore{i}}
\newcommand{\oscorej}{\oscore{j}}
\newcommand{\gscorename}{\mathtt{s_g}}
\newcommand{\gscore}[1]{\mathtt{s}^{(#1)}_{\mathtt{g}}}

\newcommand{\gscorej}{\gscore{j}}

\newcommand{\cscore}[1]{\mathtt{s}^{(#1)}_{\mathtt{c}}}

\newcommand{\cscorej}{\cscore{j}}
\newcommand{\maxiter}{b}
\newcommand{\kneeradius}{\lradiusname_{\mathtt{kr}}}
\newcommand{\kneedlename}{\mathtt{kneedle}}
\newcommand{\kneedle}[1]{\kneedlename(#1)}
\newcommand{\pointout}{\mathbf{P}}

\newcommand{\cspace}{\hspace{1\tabcolsep}}

\begin{document}

\hyphenation{Definition}
\hyphenation{Definitions}

\newcommand\relatedversion{}

\title{\Large \method: Catching \& Calling Outliers by Type\relatedversion}
\author{Guilherme D. F. Silva\thanks{University of São Paulo, Brazil. {guilhermedfs@gmail.com}}
\and Leman Akoglu\thanks{Carnegie Mellon University, USA. {lakoglu@andrew.cmu.edu}}
\and Robson L. F. Cordeiro\thanks{University of São Paulo, Brazil. {robson@icmc.usp.br}}
}

\date{}

\maketitle


\fancyfoot[R]{\scriptsize{Copyright \textcopyright\ 2022 by SIAM\\
Unauthorized reproduction of this article is prohibited}}





\begin{abstract} \small\baselineskip=9pt 
Given an unlabeled dataset, wherein we have access \textit{only} to pairwise similarities (or distances), how can we effectively (1)~detect outliers, and (2)~annotate/tag the outliers by \textit{type}?
Outlier detection has a large literature, yet we find a key gap in the field: to our knowledge, no existing work addresses the \textit{outlier annotation problem}.
Outliers are broadly classified into $3$ types, representing 
distinct patterns that could be valuable to analysts:
\begin{enumerate*}[label=(\alph*)]
    \item \textit{global} outliers are severe yet isolate cases that do not repeat, e.g., a data collection error;
    \item \textit{local} outliers diverge from their peers within a context, e.g., a particularly short basketball player; and
    \item \textit{collective} outliers are isolated microclusters that may indicate coalition or repetitions, e.g., frauds that exploit the same loophole.
\end{enumerate*}
This paper presents \method: a novel and effective outlier detector that annotates outliers by type. It is parameter-free and scalable, besides working only with pairwise similarities (or distances) when it is needed. 
We show that \method\ achieves on par or significantly better performance than  state-of-the-art detectors when spotting outliers regardless of their type. It is also highly effective in annotating outliers of particular types, a task that none of the baselines can perform.

\end{abstract}

\section{Introduction} \label{sec:intro}


This work considers a novel outlier mining problem: Given an unlabeled point-cloud dataset (wherein we may have access {only} to the pairwise similarities, or distances, between points---as opposed to feature representations), we propose \method to (1) effectively detect (``catch'') the outliers, and also (2) tag/annotate (``call'') the outliers by their \textit{type}.

\begin{figure}[!t]
\begin{tabular}{cc}
        \includegraphics[width=0.375\linewidth]{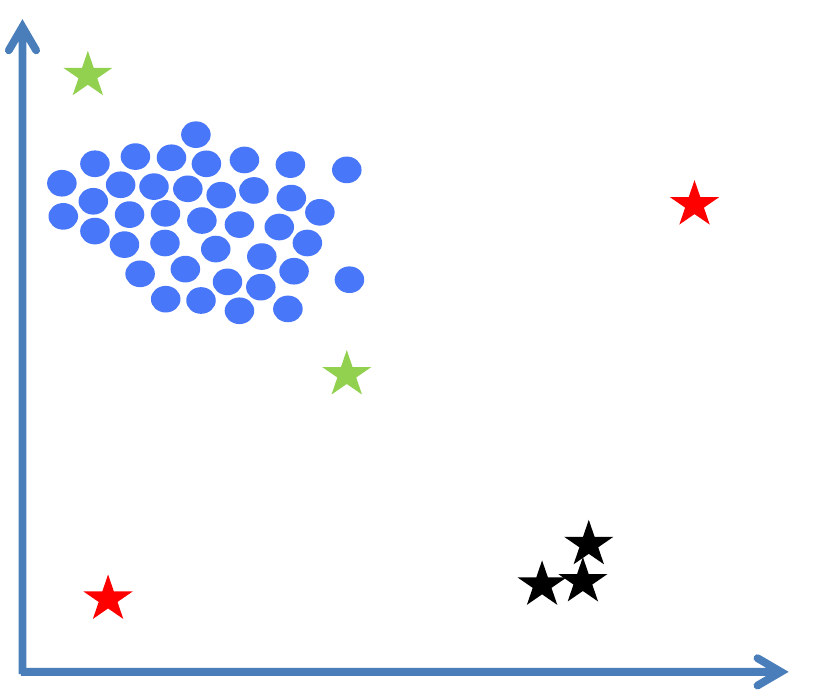}
       &
        \includegraphics[width=0.575\linewidth]{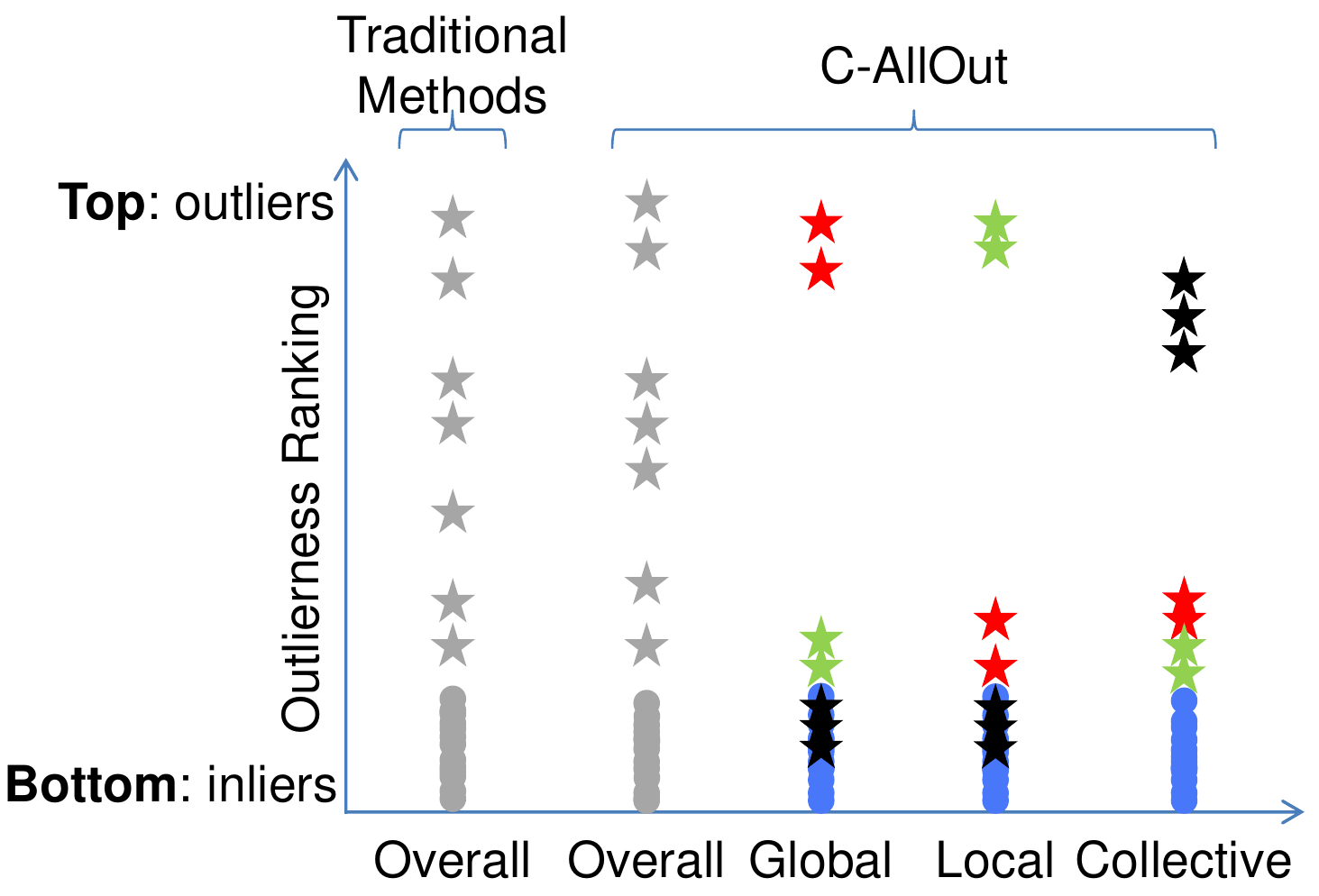} \\
        (a) example input & (b) example output
        \end{tabular}
        \vspace{-0.1in}
    \caption{\small{(a) Example toy data with inliers (circles) and 3 types of outliers (stars): global (red), local (green), and collective (black);  (b)
    Existing methods provide an overall outlier ranking, regardless of type.
    \method also provides separate rankings for outlierness by type.
    }}
    \label{fig:crown_input_output}
    \vspace{-0.25in}
\end{figure}

Outlier mining is a useful task for many applications, for which a vast body of detection techniques exists
\cite{han2012outlier}.
Despite this popularity, we identify a key gap in the literature: methods that not only can detect, but also \textit{annotate the type of the outliers detected}. 
Point-cloud datasets exhibit three main types of outliers: global, local (or contextual), and collective (or clustered), as illustrated in {Fig. \ref{fig:crown_input_output}(a).}
While several existing methods \cite{inne,lof,kriegel2008angle,outrank,loci} may potentially detect all three types of outliers, they do not label or annotate them \textit{explicitly}. The typical output of the existing detectors in the literature is either a binary label (outlier/inlier), or an overall outlierness score per point---not an ``outlier label'', i.e. an annotation reflecting its type. In fact, (to our knowledge) there is no existing detection algorithm that can annotate outliers of all types. We found SRA \cite{nian2016auto} to be the closest attempt, 
however, it provides a \textit{single} ranking along with a binary flag: scattered vs. clustered. This does not adequately address the task (See related work in \S\ref{sec:related}), and performs poorly in experiments (\S\ref{sec:experiments}). 

As outliers are key indicators of faults, inefficiencies, malicious activities, etc. in various systems, outlier annotations could provide valuable information to analysts for sensemaking, triage and troubleshooting. Global outliers can be interpreted as the most severe yet isolate cases that do not repeat; e.g., a human mistake that occurred while typing the value of an attribute, say to register value $10$ instead of $1.0$.
Local ones are outliers with respect to a specific context, like a particularly short basketball player that could be seen as an inlier among non-players, or a hot day in the winter that may be interpreted as usual within days from other seasons.
Finally, collective outliers may be critical for settings such as fraud detection, indicating coalition or repetitions, as compared to isolate `one-off' outliers. For example, they could be attacks that exploit the same loophole in cyber systems, or bots under the same command-control.



Annotating outliers by \textit{type} -- as global, local, or collective -- is a more general, and a strictly harder problem than mere detection.
Even though it is a categorization task, it remains an \textit{unsupervised} one, provided that the detection task/dataset itself does not contain any labels. 
While one may think of segmenting the detected outliers by a given detector into various types {\em post-hoc} (i.e., proceeding detection), how to do so is not obvious.
Intuitively, a post-hoc annotation procedure would require knowledge about the clusters of similar points within the entire dataset; this would both define the potential contexts for local outliers and allow the identification of microclusters formed by points with high outlierness scores, so to treat them as collective outliers.
However, clustering itself is a hard task; most of the existing algorithms require user-defined parameters, like the number of clusters $k$ in $k$-means or the minimum density ${MinPts}$ in DBSCAN \cite{ester1996density}, and still struggle with cases where the clusters have arbitrary shapes and/or distinct densities, or even when the data has no clusters at all.
These are probably the main reasons why we do not see any attempts of post-hoc annotation of outliers in the literature.



We remark that
(to the best of our knowledge) the problem of annotating outliers has not been addressed by any prior work. Notably our solution is \textit{self-contained}, that is, we do not treat this problem as a separate, post-hoc attempt, but rather propose a new outlier detection approach, called \method, that can {\em simultaneously detect  and annotate} all three types of outliers in point-cloud datasets.
Specifically, \method\ builds on the Slim-tree~\cite{slimtree}, an efficient data indexing structure with log-linear time complexity, that queries only $O(n\log n)$ pairwise similarities between points for construction. Based on this data structure, we introduce novel measures that are capable of effectively quantifying the outlierness of each point by type; as global, local or collective outlier.

As a by-product, our work can handle outlier mining settings for which \textit{only pairwise similarities can be accessed}, i.e., data points do not reside in an explicit feature space, but the similarity between any pair of points can be queried.
This setting may arise in domains with complex objects, such as galaxies, cellular tissues, etc., where providing (relative) similarities between objects may be more intuitive or convenient for domain experts than explicitly denoting them with specific feature values.
Privacy-sensitive domains, such as medicine and finance, are other settings where pairwise comparison (of patients, customers, etc.) may be deemed less risky than revealing their explicit feature values.
While our work concerns this setting, we are not limited by it; for traditional settings
where points lie in an explicit feature space, pairwise similarities can easily be computed  
provided a meaningful similarity function. 




We summarize our main contributions  as follows.

\cbit 

\item{\bf New outlier mining algorithm:} We introduce \method, a novel ``catch-n-call'' algorithm that can not only effectively detect but also annotate the outliers by type. To our knowledge, no existing work addresses the outlier annotation problem.

\item{\bf Annotating outliers:}  
Given a point-cloud, \method 
creates four separate rankings
(See {Fig. \ref{fig:crown_input_output}(b)}): an overall ranking, as well as rankings that reflect global, local and collective outlierness; all of which are estimated efficiently from the same underlying data structure  \method builds on.

\item{\bf Other desirable properties:} 
Besides providing outlier annotations, \method exhibits several other desirable properties for practical use. First, it is parameter-free and requires no user input.
Moreover it is scalable, with log-linear time complexity 
for model construction and outlier scoring.
Finally, it can handle scenarios where only
pairwise similarities between points are available (but not explicit feature representations), while not being restricted to this setting.

\item{\bf Effectiveness \& New  benchmark datasets:} We evaluate \method against  state-of-the-art detectors 
where it achieves on par or significantly better performance on standard benchmark datasets.  On additional carefully designed testbeds with annotated outliers, we also show the efficacy of \method at outlier annotation---a task none of the baselines can address. On comparable results (only), \method is superior to SRA \cite{nian2016auto}. 
\ceit 

To foster future work on the outlier annotation problem and for reproducibility, we open source \method, all benchmark datasets with annotated outliers and data generators, at \url{https://bit.ly/3iUVwtM}.




\section{Related Work} \label{sec:related}

Outlier mining has a  large literature on detection algorithms as it finds applications in numerous real world domains \cite{books/sp/Aggarwal2013}. 
Our work concerns algorithms for point-cloud data, where existing detectors can be categorized in various ways; e.g. based on  
employed measures 
such as 
distance-, density-, angle-, depth- etc.,  
or by the modeling approach 
such as 
statistical-modeling, clustering,
ensemble, subspace, etc. among many others \cite{han2012outlier}.

Here we discuss detection algorithms under a different light, i.e. with respect to \textit{outlier types}, 
as our work  focuses on detecting \textit{and annotating} outliers by type.
Namely, point-cloud datasets may contain three types of outliers: global (or gross), local (or contextual/conditional), and collective (or clustered/group).

Most detection algorithms assume outliers to be scattered, isolated points that are far from the majority \cite{knorr1998algorithms,kriegel2008angle,knn,scholkopf2000support}.
Such outliers are referred to as global outliers or gross anomalies.
A second class of detectors target contextual outliers, defined as points that are different under a certain context \cite{liang2016robust,conf/pkdd/MeghanathPA18}. These 
can be seen as 
local outliers that deviate locally \cite{lof,dang2013local,kriegel2009loop}. A third category of methods focus on  collective outliers, defined as
groups of points that collectively deviate significantly from the rest, even if the
individual members may not be outliers \cite{han2012outlier}.

Note that specialized techniques  for outliers of certain types often also detect other types, without differentiating them by type. For instance, LOF  \cite{lof} for local outlier detection also gives large, if not larger, scores to global outliers. 
Similarly,  $k$NN  \cite{knn} would give large outlier scores to collective outlier points provided $k$ is larger than the size of the collective group. In fact, all of these methods are evaluated on classical benchmark datasets w.r.t. \textit{overall} detection performance. 
On the other hand, there are techniques specifically designed to detect all three types of outliers \cite{inne,lof,kriegel2008angle,outrank,loci}. 
However none of these approaches provides any means for annotation.
{Our proposed \method is designed exactly to fill this gap in the literature.}

The closest work to ours is the SRA algorithm \cite{nian2016auto} with key distinctions. Unlike \method, it provides only a \textit{single} ranking, along with a binary flag to indicate if the ranking reflects clustered or scattered outlierness (not both).  Moreover, for the latter, it does not distinguish between global and local outliers.

\hide{
While specialized techniques have been designed to detect outliers of certain types, as discussed above, it is not to conclude that they will {\em only} detect those types. 
For instance, LOF algorithm \cite{lof} for local outlier detection also gives large, if not larger, scores to global outliers in a dataset.
Similarly,  $k$NN algorithm \cite{knn} would give large outlier scores to collective outlier points provided $k$ is larger than the size of the collective group.
Thus existing methods, while being able to detect certain types of outliers, often also detect other types in the process, without differentiating them by type (recall that no existing method provides outlier annotations). In fact, all of these methods are evaluated on classical benchmark datasets w.r.t. \textit{overall} detection performance. 
Even if these specialized methods detected only the types of outliers they were originally designed for, there would be no single all-around (i.e. ``one-stop shop'') method to detect all and any type of outliers \textit{with} annotations.
{Our proposed \method is designed exactly to fill this gap in the literature.}
}

Besides annotating outliers, \method exhibits 
other desirable properties, 
namely it is parameter-free, scalable, and uses solely pairwise similarities.
In comparison, most
outlier detectors exhibit user-defined (hyper)parameters (HPs), 
e.g. number of nearest neighbors (NNs) $k$ for NN-based methods \cite{lof,knn}, distance threshold \cite{knorr1998algorithms}, 
fraction of outliers \cite{knorr1998algorithms,scholkopf2000support}, 
etc. 
It is understood that most detectors are quite sensitive to their HP choices 
\cite{goldstein2016comparative}, however, it is unclear how to set them since hold-out/validation data with ground-truth labels simply does not exist for unsupervised detection. 
Moreover, the most prevalent techniques are distance- or density-based and rely on NN distances which are either expensive to compute or need to be approximated \cite{lof,knorr1998algorithms,loci,knn,scholkopf2000support}.  
Finally, as we motivated in the previous section, various detection algorithms that work 
in an explicit feature space are not suitable for settings where only the pairwise similarities can be provided e.g. for privacy or expert convenience reasons. 
Unfavorably, 
it is the $k$-NN based methods that can readily handle pairwise similarities, which however are computationally demanding and sensitive to the choice of their HP(s).
We remark that \method is free of these practical challenges, while addressing the novel outlier annotation problem for the first time.




\section{Problem Statement \& Preliminaries} \label{sec:back}

Considering (to the best of our knowledge) \method is the first solution to outlier annotation, we first  formalize the problem. We also 
describe Slim-trees; the  underlying data structure that \method builds on.

\subsection{Problem Statement~}
\vspace{-0.05in}

\begin{Definition}
[\textbf{Outlier Annotation Problem}]{Given a point-cloud  dataset $\dataset=\{ \point{1}, \dots, \point{n} \}$ in which only the distances, or similarities, $\distij$ between any two objects $\pointi$ and $\pointj$ can be accessed; 
Provide four types of ranking of the objects by outlierness:
1) $\orank$, an \textit{overall} ranking of outliers regardless of their type; as well as 
a ranking by 2) \textit{global}, 3) \textit{local}, and 4) \textit{collective} outlierness, respectively denoted by 
$\grank$, 
$\lrank$, and
$\crank$.
\label{definition:problem}
}
\vspace{-0.05in}
\end{Definition}

\begin{Definition}[\textbf{Ranking of Outliers}]{
A ranking of outliers $\rank=( r_1, r_2, \dots, r_n )$ is a permutation of the set $\{1, 2, \dots, n\}$. It defines an ordering of the instances in a dataset $\dataset=\{ \point{1}, \dots, \point{n} \}$ by separating the outliers
$\{\point{r_1}, \point{r_2}, \dots, \point{r_m}\}$ from the inliers $\{\point{r_{m+1}},\point{r_{m+2}}, \dots,$ $\point{r_n}\}$, where $m \ll n$ is the number of outliers in $\dataset$.
\label{definition:ranking}
}\end{Definition}



\subsection{Slim-tree~} \label{sec:slimtree}


Slim-tree~\cite{slimtree} is a tree-based data structure.
It relies solely on distances (or similarities) $\distij$ between pairs of objects $\pointi, \pointj \in \dataset$, and can be built in $O(n \log n)$ time using any metric distance function for point-cloud data in a feature space.
The distances can be either given to the tree without using features, or computed directly from them.

A tree~$\tree$ has nodes organized hierarchically into levels.
As shown in Def.ns~\ref{definition:root}, \ref{definition:leaf}, \ref{definition:anc} and~\ref{definition:rep},
tree $\tree$ has always one single root node $\rootnode$.
Root $\rootnode$ stores into a set $\rootpoints$ at most $c \in \mathbb{Z}^+$ objects from dataset $\dataset$, where $c$ is the maximum node capacity.
Each object $\repany \in \rootpoints$ is said to be the representative of a child $\node$ of the root; thus, root $\rootnode$ has $|\rootpoints|$ children.
Node $\node$ also follows the same structure: it stores into a set $\bN$ at most $c$ objects from $\dataset$.
It is always true that $\rootpoints \cap \bN = \{ \repany \}$, which means that $\repany$ is the one and only object stored redundantly both in $\node$ and in its direct ancestor $\antany=\rootnode$.
Note that $\repany$ is always nearby the \textit{most centralized object} in $\bN$.
This organization continues recursively with node $\node$ having $|\bN|$ child nodes, each of which also having children of its own, until reaching a leaf node $\leafname$.
Leaf $\leafname$ has the same structure of any other node, except that it has no children.
Note that every object $\pointi \in \dataset$ is stored into the set $\leafipoints$ of one and only one leaf $\leafi$.
\vspace{-0.05in}
\begin{Definition}[Root] {
The root node $\rootnode$ of a tree $\tree$ built for a dataset $\dataset$
is the single node existing at the highest level of $\tree$.
The set of (root) objects stored in $\rootnode$ is denoted by $\rootpoints$
, where $\rootpoints \subseteq \dataset$. 
\label{definition:root}
}
\vspace{-0.05in}
\end{Definition}

\begin{Definition}[Leaf] {
The leaf node of an object of interest $\pointi \in \dataset$ 
in a tree $\tree$ is the only node $\leafi$ existing at the lowest level of $\tree$ such that $\pointi \in \leafipoints$, where $\leafipoints \subseteq \dataset$ is the set of objects stored in $\leafi$.
\label{definition:leaf}
}\end{Definition}

\begin{figure}[!th]
\begin{tabular}{cc}        \includegraphics[width=0.44\linewidth]{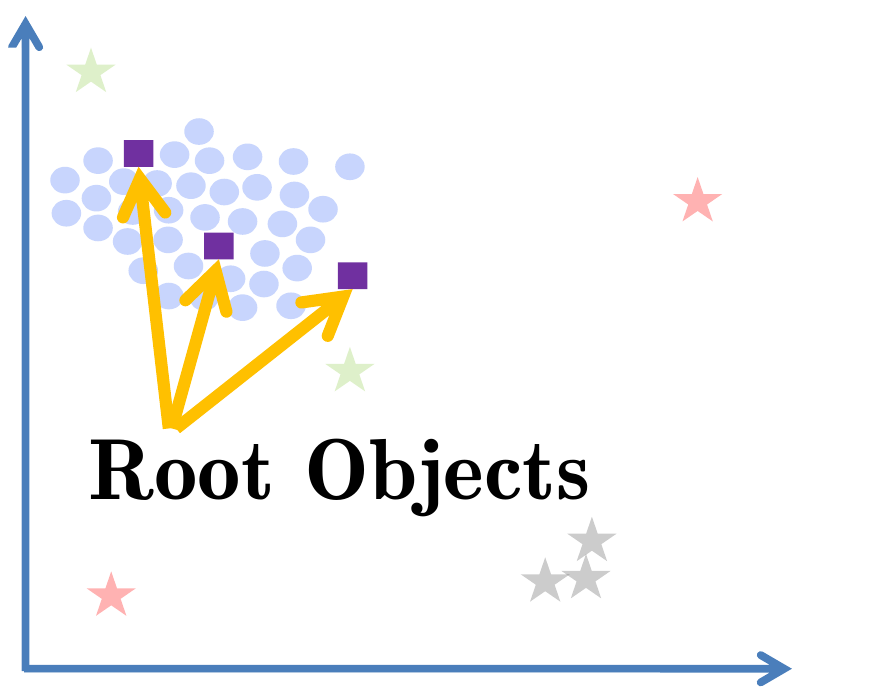}
       &
        \includegraphics[width=0.44\linewidth]{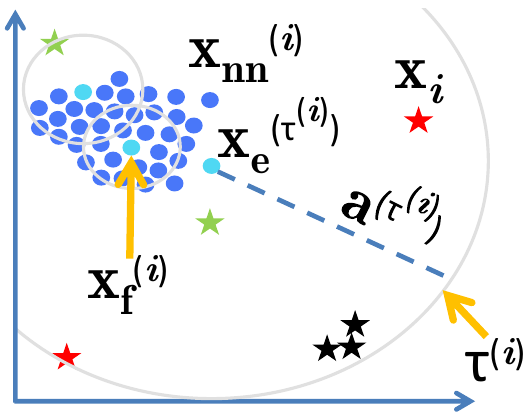} \\
        (a) root node & (b) leaf nodes
        \end{tabular}
        \vspace{-0.1in}
    \caption{Tree $\tree$ for the toy data in Fig. \ref{fig:crown_input_output}(a) of \S\ref{sec:intro}. 
    }
    \label{fig:tree}
    \vspace{-0.2in}
\end{figure}

\vspace{-0.05in}
\begin{Definition}[Direct Ancestor] {
The direct ancestor node $\antany$ of a node $\node \neq \rootnode$
within a tree $\tree$ 
is the node that directly points to $\node$ from the previous level of $\tree$. 
Note that $\ant{\rootnode}$ is undefined for root node $\rootnode$.
\label{definition:anc}
}
\vspace{-0.05in}
\end{Definition}

\vspace{-0.05in}
\begin{Definition}[Representative] {
The representative object $\repany \in \dataset$ of a node $\node \neq \rootnode$ in a tree $\tree$ built for a dataset $\dataset$ 
is the only object both in $\node$ and in $\antany$.
Note that $\rep{\rootnode}$ is undefined for root node $\rootnode$.
\label{definition:rep}
}
\vspace{-0.05in}
\end{Definition}

Fig.~\ref{fig:tree} shows a tree $\tree$ built for the toy data in Fig. \ref{fig:crown_input_output}(a) of \S\ref{sec:intro}. 
Each of the $3$ root objects in (a) represents one leaf node in (b).
One of them is leaf $\leafi$ with radius $\lradiusi$. Object $\pointi$ is in $\leafi$;
its nearest neighbor is $\nni$.

\section{Proposed Method} \label{sec:method}


\subsection{\method in a Nutshell~}
The main idea behind our method is to tackle the outlier annotation problem by leveraging the information present in a tree-based indexing structure, namely the Slim-tree.
Particularly, \method takes advantage of the ability of this data structure to group together similar instances and keep apart distinct instances, with varying ``degrees'' of similarity being captured by means of distinct tree levels.
The pseudocode of \method is in Algo.~\ref{algorithm:main}.

In line with the problem statement in Def.n~\ref{definition:problem}, \method receives a dataset $\dataset$ as input, and  produces four rankings of outliers $\orank$, $\grank$, $\lrank$ and $\crank$, respectively reflecting outlierness overall, as well as global, local and collective outlierness.
A parameter $\maxiter \in \mathbb{Z}^+$ is also received as input, for which we provide a reasonable default value $\maxiter=10$ to free \method from user-defined parameters.
The default value is based on a comprehensive empirical evaluation, see  \S\ref{ssec:refine}.

In a nutshell, \method has two main phases.
It begins in Phase~$1$ with a data-driven procedure that automatically creates and fine-tunes the base tree structure; see Lines~$1$-$9$ in Algo.~\ref{algorithm:main}. In this phase, we perform 
an iterative refinement of the tree which is controlled both by the overall ranking produced from preliminary trees, as well as by the maximum number of iterations $\maxiter$.
Once a fine-tuned tree is obtained, Phase~$2$ takes place; see Lines~$10$-$12$ in Algo.~\ref{algorithm:main}. It extracts from the refined tree the information required to rank outliers by type, and produces the final output.
The next two subsections give details on each of these phases. 


\begin{algorithm}[!t]
{\small{
\caption{\method}
\label{algorithm:main}
\textbf{Input} Dataset $\dataset=\{ \point{1}, \point{2}, \dots \point{n} \}$;
maximum number of iterations (default: $\maxiter=10$)\\
\textbf{Output} Overall ranking $\orank$; 
Global ranking $\grank$; 
Local ranking $\lrank$; 
Collective ranking $\crank$.
\begin{algorithmic}[1]
\LineComment{\textcolor{blue}{Phase $1$: building and refining the tree}}
\State $\rank \gets (1, 2, \dots n)$;
\Comment{given order of objects}
\For{$i = 1, 2, \dots \maxiter$}
    \State $\tree \gets $ {\sc CreateTree($\dataset$, $\rank$)};
    \State $\orank \gets $ {\sc Overall($\dataset$, $\tree$)}; \Comment{ Ranking~\ref{definition:overall} (\S\ref{sec:overall})}
    \If{$\orank == \rank$}
        \State \textbf{break};
    \EndIf
    \State $\rank \gets \orank$;
\EndFor
\LineComment{\textcolor{blue}{Phase $2$: outlier-type focused ranking of objects}}
\State $\grank \gets $ {\sc Global($\dataset$, $\tree$)}; \Comment{ Ranking~\ref{definition:global} (\S\ref{sec:global})}

\State $\crank \gets $ {\sc Collective($\dataset$, $\tree$)}; \Comment{ Ranking~\ref{definition:collective} (\S\ref{sec:collective})}
\State $\lrank \gets $ {\sc  Local($\dataset$, $\tree$)}; \Comment{ Ranking~\ref{definition:local} (\S\ref{sec:local})}

\State \Return $\orank, \grank, \lrank, \crank$;
\end{algorithmic}
}}
\end{algorithm}
\setlength{\textfloatsep}{0.15in}

\vspace{-0.05in}
\subsection{Building and Refining the Tree}


We build and refine a Slim tree as the base to both rank and annotate the outliers. 
For construction, the tree relies solely on pairwise distances (or similarities), and
can be built using any metric distance function for point-cloud data in a feature space.
This property also makes \method  applicable to settings where only distances between objects are available, but not explicit features.
Importantly, not all but only $O(n\log n)$ pairwise distances are required for construction, making it scalable to large datasets.




In the following two subsections, we (1) describe how 
to use the tree to create an overall outlierness ranking of the objects, which
 is then used to (2) improve/re-construct the tree structure by ensuring that inliers are more likely to be inserted early on.

\subsubsection{Overall Ranking of Outliers}\label{sec:overall}


Our overall ranking $\orank$ 
is based on two structural components of the tree in 
Def.ns~\ref{definition:croot} and~\ref{definition:frep} which are presented first.


\vspace{-0.05in}
\begin{Definition}[Closest Root Object]{
The closest root object $\crooti$ 
of an object $\pointi \in \dataset$ 
inside a tree~$\tree$ is given by
$\crooti := \{\pointj \in \rootpoints \;|\; \arg\min_j \distij\}$.
\label{definition:croot}
}
\vspace{-0.05in}
\end{Definition}

\begin{Definition}[Foreign Representative]{\sloppy{The foreign representative $\frepi$ 
of an object  $\pointi \in \dataset$ 
indexed by a tree $\tree$ is given by $\frepi := \{\pointj \in \bE \ | \ \arg\min_j \distij\}$, where
$\bE = \{ \repany \;|\; \node \in \tree \ \land \ \antany = \anti \}$.} \label{definition:frep}
}\end{Definition}


\begin{ranking}[\textbf{Overall Ranking}]{
The overall ranking of outliers $\orank = ( r_1, r_2, \dots, r_n )$ for a dataset $\dataset=\{ \point{1}, \dots, \point{n} \}$ indexed by a tree $\tree$ is 
the ranking in which $\oscore{r_i} \geq \oscore{r_{i+1}} \ \forall i \in \{ 1,2,\dots, n-1\}$, where
\vspace{-0.1in}
\begin{equation}
\label{eq:overall_score}
 \oscorej = \dist{\pointj}{\crootj} \ \cdot \ \dist{\pointj}{\frepj}
\end{equation}
\label{definition:overall}
\vspace{-0.3in}
}\end{ranking}
The main idea 
is to leverage the set of objects $\rootpoints$ that are stored in the root node~$\rootnode$ of the tree $\tree$ built from the input dataset $\dataset$, 
by 
recognizing that $\rootpoints$ is in fact a small and data-driven sample (i.e., subset of objects from $\dataset$)  containing only the objects from $\dataset$ that best summarize the main clusters of inliers.
As described in \S\ref{sec:slimtree}, the leaf nodes of $\tree$ store all the objects in $\dataset$, with each leaf having only objects that are very similar to each other.
Distinctly, an internal node solely has one object from each one of its child nodes: the most centralized object in the child.
It means that any internal node is essentially a cluster of clusters.
Notably, the information in the tree closely resembles the output of a hierarchical clustering algorithm.

As a consequence, it is reasonable to expect that an object $\pointi \in \dataset$ that clearly belongs to a cluster of inliers tends to be the one selected as the representative $\repi$ of its leaf node $\leafi$, and then again selected to be the representative 
of the direct ancestor $\anti$ of leaf $\leafi$, and still continue being selected as a representative up to the root node $\rootnode$, thus becoming a member of set~$\rootpoints$.
This line of thought lead us to believe that each object in $\rootpoints$ represents one of the clusters of inliers  in $\dataset$, where the larger is the cluster, the more representatives it has\footnote{Note that $\rootpoints$ typically has a few hundreds of objects, so we expect $k \ll |\rootpoints|$, where $k$ is the number of inlier clusters in $\dataset$.}.
Following Def.n~\ref{definition:croot}, let $\crooti$ be the object in $\rootpoints$ that is the closest to an object of interest~$\pointi$. For short, we simply say that $\crooti$ is the closest root object of $\pointi$.
Note that it helps us to distinguish outliers from inliers: if $\pointi$ is an inlier, then it tends to be closer to $\crooti$ than it would be if it were an outlier.

Nonetheless, some small clusters of inliers may be underrepresented in the root node.
In this particular case, the distance to the closest root object becomes less helpful.
Fortunately, there also exists the tendency that a cluster of inliers requires more than one leaf node to be stored, since any tiny cluster that fits entirely into a single leaf is more likely to contain collective outliers than inliers.
This tendency also helps us to distinguish outliers from inliers: if an object $\pointi$ is close to at least one leaf other than leaf $\leafi$, it tends to be an inlier; otherwise, $\pointi$ is probably an outlier.
Following Def.n~\ref{definition:frep}, we efficiently take advantage of this idea through the concept of foreign representative~$\frepi$ of an object $\pointi$.
The overall ranking of outliers $\orank$ shown in Ranking~\ref{definition:overall} builds on the aforementioned concepts by using a new score $\oscorei$ as a base.
It capitalizes on the proximity of an object $\pointi$ both to its closest root object~$\crooti$ and to its foreign representative~$\frepi$ to accurately and efficiently rank outliers of any type.

\vspace{-0.05in}
\subsubsection{Sequential Refinement of Rankings}
\label{ssec:refine}
The tree structure that we leverage is built by a sequential insertion of the objects, and has its own heuristics to efficiently avoid unfortunate selections of node representatives that nonetheless may occur depending on the order in which the objects are given for insertion.
Next we describe
how to improve the initial tree created by following a given order of objects, say, random insertion, into a fine-tuned tree created from a better ordering of the objects---making \method robust to any order of objects given for a dataset $\dataset$.


As can be seen in Lines~$1$-$9$ of Algo.~\ref{algorithm:main}, we propose to achieve this goal by performing a sequential refinement of rankings. 
Specifically, an initial tree $\tree$ is created by following a given order of objects $\rank$; then, $\tree$ is iteratively refined by exploiting the overall ranking $\orank$ produced from it. 
Note that function {\sc CreateTree($\dataset$, $\rank$)} in Line~$3$ of Algo.~\ref{algorithm:main} uses the \textit{reverse} order of $\rank$ to insert objects in $\tree$. 
The idea behind this procedure is to insert clear inliers first in $\tree$, so as to avoid the selection of an outlier to be a node representative.
As we showcase empirically in  Suppl. \S\ref{sec:sup_refinement}, only a handful of iterations is sufficient to refine the insertion order (and as a result the outlier ranking), as most outliers are typically detected even after the first round.

\vspace{-0.05in}
\subsection{Outlier-type Focused Ranking of Objects}\label{sec:ranktypes}
Moving further from an overall ranking, we next describe
how \method ranks outliers of each particular type; focusing 
first on global outliers, and then the collective and local outlier ranking.

\vspace{-0.05in}
\subsubsection{Global Ranking of Outliers} \label{sec:global}

Here we build on
the fact that the overall ranking 
by Eq. \eqref{eq:overall_score}
tends to rank global and collective outliers ahead of the other objects.
Recall that the overall ranking $\orank$ ranks each object according to its distance to the closest cluster of inliers.
Both global and collective outliers tend to be far away from any of such clusters;
distinctly, local outliers are close to at least one cluster, while the inliers themselves are even closer.
Consequently, we expect that  $\orank = ( r_1, r_2, \dots, r_n )$ sorts any two objects $\point{r_i}, \point{r_j} \in \dataset$ such that $i < j$ only when $\point{r_i}$ is either a global or a collective outlier, and $\point{r_j}$ is not.
Thus, the main challenge here is to differentiate the global outliers from the collective ones.

Fortunately, these two types of outliers can be distinguished: collective outliers have close neighbors; global ones do not.
Building on Def.ns~\ref{definition:nn} and \ref{definition:nnd},  
we form the ranking $\grank$ of global outliers by combining
\begin{enumerate*}[label=(\alph*)]
\item 
the score $\oscorei$ of an object~$\pointi$, which is the core of our overall ranking; with
\item 
the normalized distance $\distnni$ between $\pointi$ and its nearest neighbor $\nni$, as shown in Eq. \eqref{eq:global}.
\end{enumerate*}
Intuitively, score $\oscorename$ separates global and collective outliers from other objects, while distance $\distnnname$ differentiates outliers of these two types.
Note that the normalization
 in 
 Eq. \eqref{eq:nnd}\footnote{Note that value $1$ in the denominator avoids division by zero.\label{note:division}}
 balances the distances according to the local neighborhood of~$\pointi$.
Also, the nearest neighbor $\nni$ is in fact an approximate one; it is efficiently obtained by comparing $\pointi$ only with the few objects in leaf $\leafi$. 
It tends to be a good approximation since our base tree is built to have each leaf storing only objects that are very similar to, i.e. nearby each other.

\begin{Definition}[Approximate Nearest Neighbor]{
The approximate nearest neighbor $\nni$ of object $\pointi \in \dataset$ in a tree~$\tree$ is $\nni := \{\pointj \in \leafipoints \;|\; \arg\min_{j \neq i} \distij\}$, where $\leafipoints$ is the set of objects stored in leaf $\leafi$.
\label{definition:nn}
}\end{Definition}

\begin{Definition}[Normalized $1$nn Distance] {
The normalized nearest neighbor distance $\distnni$ of an object $\pointi \in \dataset$ indexed in a tree~$\tree$ is defined as
\vspace{-0.05in}
\begin{equation}
\distnni = \frac{\dist{\pointi}{\nni}}{1 + \dist{\repi}{\nn{\leafi}}}
\label{eq:nnd}
\vspace{-0.1in}
\end{equation}
where $\nn{\leafi}~:=~\nnj$ s.t.  
$\pointj$ is equivalent to $\repi$.
\label{definition:nnd}
}\end{Definition}

\begin{ranking}[\textbf{Global Ranking}]{
The global ranking $\grank = ( r_1, \dots, r_n )$ for dataset $\dataset$ is 
the ranking in which $\gscore{r_i} \geq \gscore{r_{i+1}} \ \forall i \in \{ 1,2,\dots, n-1\}$, where
\vspace{-0.05in}
\begin{equation}
 \gscorej = \oscorej \ \cdot \ \distnnj
 \label{eq:global}
\end{equation}
\label{definition:global}
\vspace{-0.3in}
}\end{ranking}





\subsubsection{Collective Ranking of Outliers} \label{sec:collective}

The ranking $\crank$ of  collective outliers
builds naturally from the ideas discussed in the previous section, 
where we exploit the tendency of global and collective outliers to have the highest scores $\oscorename$,
and then ``amplify'' each object $\pointi$ that is a global outlier by multiplying $\oscorei$ by the nearest neighbor distance $\distnni$.
In a similar fashion, we bump up collective outliers simply by replacing the multiplication with its reverse operation division\footref{note:division}, as
presented in Eq. ~\eqref{eq:collective}.

\begin{ranking}[\textbf{Collective Ranking}]{
The collective ranking $\crank = (r_1,  \dots, r_n )$ for dataset $\dataset$ is the ranking in which $\cscore{r_i} \geq \cscore{r_{i+1}} \ \forall i \in \{ 1,\dots, n-1\}$, where
\vspace{-3mm}
\begin{equation}
 \cscorej = \frac{\oscorej}{1+\distnnj}
 \label{eq:collective}
\end{equation}
\label{definition:collective}
\vspace{-0.2in}
}\end{ranking}





\subsubsection{Local Ranking of Outliers} \label{sec:local}




The local ranking of outliers $\lrank$ is formally given in Ranking~\ref{definition:local} below, which builds on Def.ns~\ref{definition:radius} and~\ref{definition:kneeradius}.

\begin{Definition}[Normalized Leaf Radius] {
The normalized leaf radius $\lradiusany$ of a leaf~$\leafname$ in a tree~$\tree$ is
\vspace{-2mm}
\begin{equation}
 \lradiusany = \frac{\dist{\repleaf}{\fnleaf}}{1 + \dist{\repleaf}{\nnleaf}}
\end{equation}
where $\fnleaf := \{\pointi \in \leafanypoints | \arg\max_i \dist{\repleaf}{\pointi}\}$
is the farthest from the leaf representative
among objects $\leafanypoints$ in leaf $\leafname$, and 
$\nnleaf~:=~\nnj$ s.t. $\pointj$ is equiv. to $\repleaf$.
\label{definition:radius}
}\end{Definition}

\begin{Definition}[Knee Radius] {
The knee radius $\kneeradius$ for a dataset~$\dataset = \{\point{1}, \dots \point{n}\}$ in a tree~$\tree$ is defined as
\vspace{-1mm}
\begin{equation}
\kneeradius = \kneedle{\llradius{1}, \dots , \llradius{|\slradiuses|}}
\vspace{-1mm}
\end{equation}
s.t. $\slradiuses=\{\lradius{\leaf{1}}, \dots, \lradius{\leaf{n}}\}$,
and
$\llradius{j} < \llradius{j+1} \ \forall j \in \{ 1,\dots, |\slradiuses|-1\}$. 
Function $\kneedlename$ refers to algorithm Kneedle~\cite{kneedle}; it takes a sorted list $\left(\llradius{1}, \dots , \llradius{|\slradiuses|}\right)$ as input and returns one of its values $\llradius{p}$.
\label{definition:kneeradius}
}\end{Definition}

\begin{ranking}[\textbf{Local Ranking}]{
The local ranking for dataset $\dataset$ is denoted by $\lrank$.
Let $\pointout = \{ \pointi \;|\; \distnni~\geq~\kneeradius \}$.
Then, $\lrank = (r_1,  \dots, r_n )$ is the ranking in which
\vspace{-1mm}
\begin{equation}
\point{r_j} \in 
\begin{cases}
    \pointout, \ \ \ \ \ \ \ \ \ \ if \ \ 1\leq j \leq |\pointout|\\
    \dataset\setminus\pointout, \ \ \ \ \ otherwise
    \vspace{-1mm}
\end{cases}
\end{equation}
where $\gscore{r_p} \leq \gscore{r_{p+1}} \ \forall p \in \{ 1,\dots, |\pointout|-1\}$ and
$\gscore{r_q}~\geq~\gscore{r_{q+1}} \ \forall q \in \{ |\pointout|+1,\dots, n-1\}$.
\label{definition:local}
}\end{ranking}

The main idea is to distinguish local outliers from other objects by using the normalized
leaf radius $\lradiusany$ of each leaf $\leafname \in \tree$.
Particularly, we expect $\lradiusany$ to be:
\begin{enumerate*}[label=(\alph*)]
\item small, if the object~$\fnleaf \in \dataset$ that defines the radius is an inlier;
\item larger, if $\fnleaf$ is a local outlier;
\item very large, if $\fnleaf$ is a global outlier;
 and
\item either very small or large\footnote{If the collective outliers are in a leaf of their own, $\lradiusany$ would be very small, and very large otherwise.}, if $\fnleaf$ is a collective outlier.
\end{enumerate*}

As a result, the vast majority of leaf radii are expected to be small, mostly coming from case (a) and some from case (d) above, with only a few large radius leaves.
Thus, when the radii in $\slradiuses=\{\lradius{\leaf{1}}, \dots, \lradius{\leaf{n}}\}$ are sorted into a list $(\llradius{1}, \dots , \llradius{|\slradiuses|})$, there must exist a ``knee'' value/radius $\kneeradius$ that clearly separates the small radii from the rest.
Intuitively, $\kneeradius$ depicts the smallest distance $\distnni$ of each object $\pointi \in \dataset$ that is a local outlier.
Our Ranking~\ref{definition:local} 
computes $\kneeradius$ using an off-the-shelf ``knee''-detector algorithm; 
then, it separates large-radii outliers in set $\pointout$, and 
reuses our global score to rank them in increasing order of $\gscorename$, effectively placing local outliers ahead of the other objects. 

\subsection{Time and Space Complexity}

\begin{lemma}[Time Complexity]{The overall time complexity of \method is $O(n \log n )$.}\label{lemma:timecomplexity}
\end{lemma}
\vspace{-0.15in}
\begin{lemma}[Space Complexity]{The overall space complexity of \method is $O(n \log n )$.}\label{lemma:spacecomplexity}
\end{lemma}
\vspace{-0.1in}
\begin{proof}
See  Suppl. \S\ref{sec:sup_time_complexity} and \S\ref{sec:sup_space_complexity}, respectively.
\end{proof}

\section{Evaluation} \label{sec:experiments}

We designed experiments to evaluate \method w.r.t. both overall detection as well as outlier annotation tasks, aiming to answer the following.
\cbit
\item[\textbf{Q1:}] {\bf Overall detection performance:} How does \method compare to state-of-the-art (SOTA) baselines on benchmark datasets w.r.t. overall  detection performance, regardless of outlier type?
\item[\textbf{Q2:}] {\bf Outlier annotation performance:} How well does \method perform on outlier annotation, that is,
w.r.t. its ranking of the outliers by type?
\ceit

\vspace{-0.05in}
\subsection{Overall Detection Performance}


\subsubsection{Setup}
\vspace{-0.05in}
First we aim to showcase that, while specializing on outlier annotation, \method is competitive in terms of overall outlier detection on standard benchmark datasets compared to SOTA baselines.
We remark that \method's overall ranking $\orank$ is used for evaluation here, regardless of outlier type.


{\bf Testbed 1.~}
We create our first testbed using classical benchmark datasets\footnote{\url{http://odds.cs.stonybrook.edu/}}$^,$\footnote{\url{http://lapad-web.icmc.usp.br/repositories/outlier-evaluation/DAMI/}\label{note:repo2}}. 
Specifically, a total of $20$ datasets were chosen at random among those with outlier percentage less than 25\%; 
a detailed list of which is in Table \ref{tab:benchmark_datasets} in Suppl. \S\ref{sec:sup_dataset_summary}.


{\bf Baselines.~} We compare against four SOTA similarity or distance-based detection algorithms; namely, LOF \cite{lof}, kNN \cite{knn}, OCSVM \cite{scholkopf2000support}, and Sim-iForest (SIF, for short) \cite{czekalski2021similarity}. 
Specifically, LOF, kNN and SIF employ Euclidean distance and OCSVM uses the linear kernel.
Note that SIF is a version of the popular iForest \cite{liu2008isolation}, constructed by using pairwise similarities only based on ideas from similarity forests \cite{sathe2017similarity}.
(See Suppl. \S\ref{sec:sup_competitors_outlier_detection})



Unlike \method, all baselines exhibit one or more hyperparameters (HPs) that are non-trivial to pick without any labels. We run each baseline using a corresponding list of HP configurations (See Table \ref{tab:sup_competitors_parameters} in Suppl. \S\ref{sec:sup_model_config}.) and report the average performance. In effect, this is equivalent to their expected performance when HPs are selected at random from the list. For the non-deterministic baseline SIF, the performance is averaged over 10 random runs.


{\bf Measures.} We evaluate the outlier ranking quality using both AUC ROC curve and AUC Precision-Recall curve. 
In addition, we apply the paired Wilcoxon signed rank test on AUC-ROC/PR results across datasets to establish statistical significance of the differences. 







\begin{table}[h]
\vspace{-0.05in}
    \scriptsize
    \caption{AUCROC performance on Testbed 1}
    \vspace{-0.1in}
    \begin{tabularx}{\columnwidth}{l|l|X|X|X|X} \hline
    Dataset & \method & LOF & kNN & OCSVM & SIF
    \\ \hline
    Annth. & 0.717 & 0.718 & \textbf{0.773} & 0.469 & 0.585
    \\
    Cardio & 0.528 & 0.535 & 0.503 & 0.380 & \textbf{0.795}
    \\
    Glass & \textbf{0.840} & 0.738 & 0.825 & 0.638 & 0.634
    \\
    Letter & 0.777 & \textbf{0.849} & 0.841 & 0.491 & 0.219
    \\
    Mammog. & 0.791 & 0.643 & \textbf{0.802} & 0.343 & 0.772
    \\
    Mnist & 0.848 & 0.662 & \textbf{0.852} & 0.408 & 0.585
    \\
    Musk & \textbf{0.954} & 0.555 & 0.795 & 0.415 & 0.919
    \\
    Optdigits & 0.465 & \textbf{0.506} & 0.437 & 0.504 & 0.372
    \\
    Pageb. & \textbf{0.903} & 0.712 & 0.877 & 0.523 & 0.852
    \\
    Pendigits & 0.848 & 0.517 & 0.813 & 0.615 & \textbf{0.879}
    \\
    Satim. & \textbf{0.998} & 0.639 & 0.961 & 0.527 & 0.894
    \\
    Shuttle & \textbf{0.986} & 0.851 & 0.953 & 0.374 & 0.765
    \\
    Shuttle2 & 0.786 & 0.517 & 0.701 & 0.647 & \textbf{0.963}
    \\
    Stamps & 0.807 & 0.643 & 0.856 & 0.530 & \textbf{0.859}
    \\
    Thyroid & 0.952 & 0.721 & \textbf{0.957} & 0.514 & 0.876
    \\
    Vowels & 0.932 & 0.885 & \textbf{0.953} & 0.549 & 0.317
    \\
    Waveform & 0.655 & 0.696 & \textbf{0.742} & 0.452 & 0.647
    \\
    WBC & 0.939 & 0.856 & \textbf{0.947} & 0.403 & 0.914
    \\
    WDBC & 0.904 & 0.831 & 0.923 & 0.474 & \textbf{0.927}
    \\
    Wilt & 0.463 & \textbf{0.635} & 0.541 & 0.296 & 0.299
    \\ \hline
    \textbf{Avg.} & 0.805 & 0.686 & \textbf{0.819} & 0.488 & 0.733
    \\ \hline
    \end{tabularx}
    \label{tab:auc_benchmark}
    \vspace{-0.25in}
\end{table}

\begin{table}[h]
    \scriptsize
    \caption{AUCPR performance on Testbed 1}
    \vspace{-0.1in}
    \begin{tabularx}{\columnwidth}{l|l|X|X|X|X} \hline
    Dataset & \method & LOF & kNN & OCSVM & SIF
    \\ \hline
    Annth. & \textbf{0.229} & 0.174 & 0.213 & 0.077 & 0.103
    \\
    Cardio & 0.322 & 0.272 & 0.312 & 0.209 & \textbf{0.582}
    \\
    Glass & 0.120 & \textbf{0.130} & 0.127 & 0.116 & 0.058
    \\
    Letter & 0.179 & \textbf{0.385} & 0.258 & 0.071 & 0.037
    \\
    Mammog. & 0.156 & 0.096 & 0.165 & 0.083 & \textbf{0.183}
    \\
    Mnist & 0.383 & 0.233 & \textbf{0.396} & 0.101 & 0.158
    \\
    Musk & \textbf{0.839} & 0.178 & 0.559 & 0.040 & 0.754
    \\
    Optdigits & 0.024 & 0.029 & 0.021 & \textbf{0.032} & 0.019
    \\
    Pageb. & \textbf{0.549} & 0.303 & 0.515 & 0.237 & 0.420
    \\
    Pendigits & \textbf{0.186} & 0.036 & 0.109 & 0.075 & 0.135
    \\
    Satim. & \textbf{0.938} & 0.050 & 0.583 & 0.108 & 0.170
    \\
    Shuttle & \textbf{0.323} & 0.220 & 0.271 & 0.011 & 0.027
    \\
    Shuttle2 & 0.290 & 0.101 & 0.182 & 0.317 & \textbf{0.776}
    \\
    Stamps & 0.255 & 0.186 & 0.283 & 0.258 & \textbf{0.289}
    \\
    Thyroid & \textbf{0.360} & 0.137 & 0.337 & 0.162 & 0.237
    \\
    Vowels & 0.347 & 0.320 & \textbf{0.477} & 0.076 & 0.021
    \\
    Waveform & 0.061 & 0.079 & \textbf{0.110} & 0.028 & 0.049
    \\
    WBC & 0.553 & 0.397 & 0.518 & 0.075 & \textbf{0.571}
    \\
    WDBC & 0.548 & 0.433 & 0.532 & 0.101 & \textbf{0.655}
    \\
    Wilt & 0.045 & \textbf{0.079} & 0.053 & 0.048 & 0.035
    \\ \hline
    \textbf{Avg.} & \textbf{0.335} & 0.192 & 0.301 & 0.111 & 0.264
    \\ \hline
    \end{tabularx}
    \label{tab:ap_benchmark}
    \vspace{-0.25in}
\end{table}

\subsubsection{Results}

Tables \ref{tab:auc_benchmark} and \ref{tab:ap_benchmark} respectively present the AUCROC and AUCPR overall detection performance results for all methods across all benchmark datasets.
In addition, Table  \ref{tab:wilcoxon} provides the $p$ values for one-sided significance tests of the differences, testing that the left-side method (\method) is better than the right-side (baseline) against the null that they are indifferent.
In terms of average AUCROC performance, \method significantly outperforms LOF, OCSVM and SIF and achieves on par with the top performer kNN (0.805 vs. 0.819) where the differences are not significant ($p\approx 0.5$).
Similar results hold for AUCPR, where \method achieves the largest average performance. The performance is comparable to kNN's (although $p$ value at 0.11 is relatively small) and significantly better than the rest.

\begin{table}[!t]
    \scriptsize
    \caption{Paired one-sided significance test (Wilcoxon): \method vs. baseline w.r.t. overall performance on Testbed 1. $p$-value in {\bf bold} if significant at 0.05.}
    \vspace{-0.1in}
    \begin{tabularx}{\columnwidth}{l|X|X} \hline
    Methods & AUCPR $p$-value & AUCROC $p$-value
    \\ \hline
    \method \textgreater \ LOF & \textbf{0.001163} & \textbf{0.001576}
    \\
    \method \textgreater \ kNN & 0.115256 & 0.492718
    \\
    \method \textgreater \ OCSVM & \textbf{0.000067} & \textbf{0.000002}
    \\
    \method \textgreater \ SIF & \textbf{0.048654} & \textbf{0.016384}
    \\ \hline
    \end{tabularx}
    \label{tab:wilcoxon}
    \vspace{-0.05in}
\end{table}

\subsection{Outlier Annotation Performance}
\vspace{-0.05in}

\subsubsection{Setup}
\vspace{-0.05in}
Next we aim to show the effectiveness of \method on annotating outliers by type, and specifically the accuracy of its type-specific rankings.
However, there exists no benchmark datasets with annotated outliers. 
To this end, we create two novel testbeds containing specific types of outliers, described as follows.

{\bf Testbed 2.~}
We start with simulating three synthetic base datasets, denoted S1, S2 and S3; consisting of Gaussian inlier clusters of varying count, size, standard deviation (std), and dimensionality. (See Table \ref{tab:synthetic_gen_parameters} in Suppl.) From each base dataset Sx, we create four versions: {Sx\_G}, {Sx\_L}, {Sx\_C}, and {Sx\_A} respectively with only global, only local, only collective, and all three types of outliers combined.
Global and local outliers are simulated from each inlier cluster by up-scaling the respective std, while each set of collective outliers are simulated from a point outlier as the mean and down-scaled std (See details in Suppl. \S\ref{sec:sup_gen_synthetic}.)




{\bf Testbed 3.} We also create a testbed 
based on Steinbuss and B{\"o}hm's ``realistic synthetic data'' generator
\cite{steinbuss2021benchmarking}.
The idea is to start with a real-world dataset containing only inliers, to which a Gaussian Mixture Model (GMM) is fit. Outliers are then simulated by up-scaling the variances of the GMM clusters. Their generator does not create collective outliers; to that end, we follow suit with Testbed 2 and simulate small-std micro-clusters centered at randomly chosen point outliers. (See details in Suppl. \ref{sec:sup_gen_realistic}.)
As the base, we use five real-world datasets (after discarding the (non-annotated) ground-truth outliers); namely  
ALOI, Glass, skin, smtp, Waveform\footref{note:repo2} with various size and dimensionality 
(See Table \ref{tab:base_datasets} in Suppl. \S\ref{sec:sup_dataset_summary}). Similar to Testbed 2, we create 4 variants (\_G, \_L, \_C, \_A) from each base dataset with various types of (annotated) outliers.




{\bf Baseline.} 
There is no baseline for annotating all three types of outliers. We found the spectral
SRA \cite{nian2016auto}, also based on pairwise similarities (i.e. Laplacian), as a close attempt (details in Suppl. \S\ref{sec:sup_competitors_outlier_annotation}). When it identifies a small cluster\footnote{SRA needs a size threshold for collective outliers; we give it an advantage by setting the threshold equal to the true size.}, it raises the flag ``clustered''  
and ``scattered'' otherwise. It provides a single ranking and does not distinguish between global and local outliers. Thus, SRA is only comparable in some cases.


\begin{table*}[!th]
    \centering
    \scriptsize
    \caption{
    AUCROC performance on Testbed 2 with annotated outliers. Note that
    in Sx\_A datasets with all three outlier types, type-specific rankings are evaluated against outliers of the corresponding type only.
    }
    \vspace{-0.1in}
    \begin{tabularx}{0.81\textwidth}{l|c @{\cspace} c|c @{\cspace} c|c @{\cspace} c|c @{\cspace} c} \hline
    Dataset & \method-$\orank$ & SRA & \method-$\lrank$ & SRA & \method-$\grank$ & SRA & \method-$\crank$ & SRA
    \\ \hline
    S1\_A & \textbf{0.997} & 0.663 & \textbf{0.999} & 0.452 & \textbf{1} & 0.851 & \textbf{0.999} & Not flagged
    \\
    S1\_L & \textbf{0.992} & 0.453 & \textbf{1} & 0.453 & - & - & - & -
    \\
    S1\_G & \textbf{1} & 0.851 & - & - & \textbf{1} & 0.851 & - & -
    \\
    S1\_C & \textbf{1} & 0.685 & - & - & - & - & \textbf{1} & Not flagged
    \\
    S2\_A & \textbf{1} & 0.703 & \textbf{0.997} & 0.511 & \textbf{1} & 0.747 & \textbf{1} & Not flagged
    \\
    S2\_L & \textbf{1} & 0.512 & \textbf{1} & 0.512 & - & - & - & -
    \\
    S2\_G & \textbf{1} & 0.744 & - & - & \textbf{1} & 0.744 & - & -
    \\
    S2\_C & \textbf{1} & 0.849 & - & - & - & - & \textbf{1} & Not flagged
    \\
    S3\_A & \textbf{1} & 0.599 & \textbf{0.998} & 0.567 & \textbf{1} & 0.729 & \textbf{1} & Not flagged
    \\
    S3\_L & \textbf{0.999} & 0.576 & \textbf{1} & 0.576 & - & - & - & -
    \\
    S3\_G & \textbf{1} & 0.730 & - & - & \textbf{1} & 0.730 & - & -
    \\
    S3\_C & \textbf{1} & 0.500 & - & - & - & - & \textbf{1} & Not flagged
    \\ \hline
    \end{tabularx}
    \label{tab:auc_synthetic}
    \vspace{-0.05in}
\end{table*}






\begin{table*}[!t]
    \centering
    \scriptsize
    \caption{AUCROC performance on Testbed 3 with annotated outliers. Note that
    in variant \_A datasets with all three outlier types, type-specific rankings are evaluated against outliers of the corresponding type only.}
    \vspace{-0.1in}
    \begin{tabularx}{0.95\textwidth}{l|c @{\cspace} c|c @{\cspace} c|c @{\cspace} c|c @{\cspace} c} \hline
    Dataset & \method-$\orank$ & SRA & \method-$\lrank$ & SRA & \method-$\grank$ & SRA & \method-$\crank$ & SRA
    \\ \hline
    ALOI\_A & \textbf{1} & 0.474 & \textbf{0.993} & Not flagged & \textbf{1} & Not flagged & \textbf{0.996} & 0.429
    \\
    ALOI\_L & \textbf{1} & 0.908 & \textbf{1} & 0.908 & - & - & - & -
    \\
    ALOI\_G & \textbf{1} & 0.400 & - & - & \textbf{1} & Not flagged & - & -
    \\
    ALOI\_C & \textbf{1} & 0.429 & - & - & - & - & \textbf{1} & 0.429
    \\
    Glass\_A & \textbf{0.966} & 0.882 & \textbf{0.886} & 0.621 & \textbf{1} & 0.938 & \textbf{1} & Not flagged
    \\
    Glass\_L & \textbf{0.892} & 0.599 & \textbf{0.938} & 0.599 & - & - & - & -
    \\
    Glass\_G & \textbf{1} & 0.985 & - & - & \textbf{1} & 0.985 & - & -
    \\
    Glass\_C & \textbf{1} & 0.951 & - & - & - & - & \textbf{1} & Not flagged
    \\
    skin\_A & \textbf{0.999} & 0.868 & \textbf{0.994} & 0.730 & \textbf{1} & 0.864 & \textbf{0.999} & Not flagged
    \\
    skin\_L & \textbf{0.998} & 0.732 & \textbf{1} & 0.732 & - & - & - & -
    \\
    skin\_G & \textbf{1} & 0.869 & - & - & \textbf{1} & 0.869 & - & -
    \\
    skin\_C & \textbf{1} & 1 & - & - & - & - & \textbf{1} & Not flagged
    \\
    smtp\_A & \textbf{0.995} & 0.986 & \textbf{0.992} & 0.951 & \textbf{1} & 0.998 & \textbf{0.997} & Not flagged
    \\
    smtp\_L & \textbf{0.986} & 0.470 & \textbf{0.998} & 0.470 & - & - & - & -
    \\
    smtp\_G & \textbf{1} & 0.999 & - & - & \textbf{1} & 0.999 & - & -
    \\
    smtp\_C & \textbf{1} & 0.713 & - & - & - & - & \textbf{1} & Not flagged
    \\
    Waveform\_A & \textbf{1} & 0.575 & \textbf{0.991} & 0.512 & \textbf{1} & 0.480 & \textbf{1} & Not flagged
    \\
    Waveform\_L & \textbf{0.999} & 0.507 & \textbf{1} & 0.507 & - & - & - & -
    \\
    Waveform\_G & \textbf{1} & 0.491 & - & - & \textbf{1} & 0.491 & - & -
    \\
    Waveform\_C & \textbf{1} & 0.728 & - & - & - & - & \textbf{1} & Not flagged
    \\ \hline
    \end{tabularx}
    \label{tab:auc_realistic}
    \vspace{-0.1in}
\end{table*}

\vspace{-0.075in}
\subsubsection{Results}
Table \ref{tab:auc_synthetic} presents AUCROC results on Testbed 2. \method achieves near-ideal overall performance, while SRA ranking is significantly poor.
\method type-specific rankings continue to achieve near-perfect performance, while SRA struggles, especially in identifying local outliers with a near-random ranking. Moreover, it fails to flag ``clustered''  on datasets that contain collective outliers.
Results are similar w.r.t. AUCPR. (See Table \ref{tab:ap_synthetic} in Suppl.)

AUCROC results on the ``realistic synthetic'' Testbed 3 are shown in Table 
  \ref{tab:auc_realistic}.
\method achieves highly competitive performance and significantly outperforms SRA in all cases. Interestingly, type-specific rankings are better than the overall ranking for those specific outliers, which shows that they are effectively specialized. SRA fails to identify/flag the collective outliers, and in some other cases with scatter (local or global outliers) wrongly raises the ``clustered'' flag.
Results w.r.t. AUCPR are similar; see Table \ref{tab:ap_realistic} in Suppl.
\section{Conclusion} \label{sec:conclusion}

We introduced \method, to our knowledge, the first approach that systematically tackles the outlier annotation problem. It provides not only an overall ranking but also a separate ranking by global, local, and collective
outlierness; all computed parameter-free and ``in-house'' (not post hoc), based on a scalable indexing tree structure.  
We showed
that \method is on par with or significantly outperforms state of the art outlier detectors on classical benchmark datasets. We also built two additional testbeds with annotated outliers and demonstrated the efficacy of \method in the annotation task.
To foster further work on this task, we open-source our testbeds and data generators along with the code for \method, at \url{https://bit.ly/3iUVwtM}.

{\scriptsize{
\bibliography{references}
\bibliographystyle{siam}
}}

\clearpage
\section{Supplementary Material for \textit{\method: Catching \& Calling Outliers by Type}}\label{sec:sup_main}

\vspace{0.1in}
\subsection{Sequential Refinement of Rankings} \label{sec:sup_refinement}

\begin{figure}[H]
    \centering
    \includegraphics[width=0.8\linewidth]{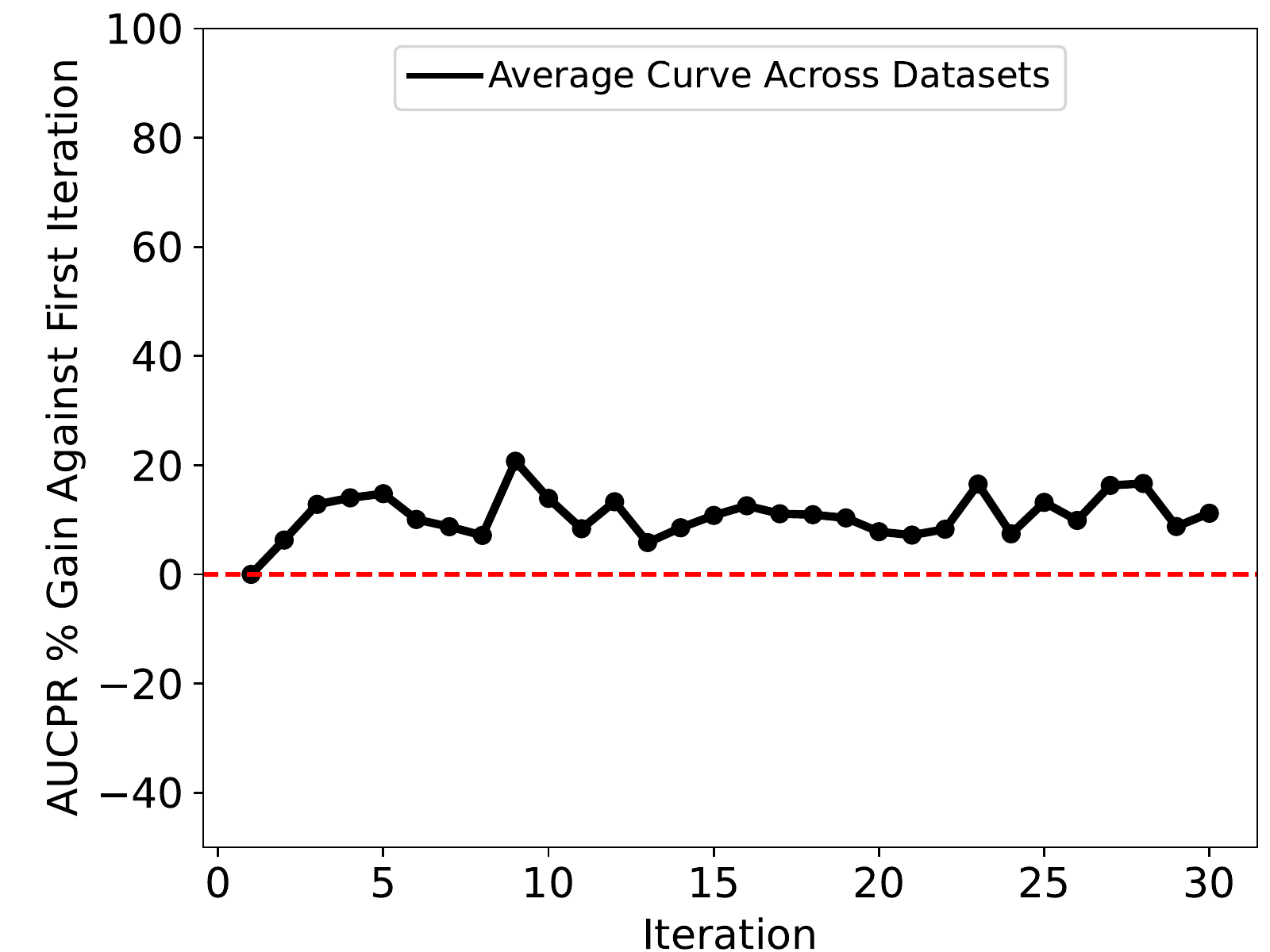}
    \caption{AUCPR performance percentage gain as data insertion order to \method is refined through iterations.}
    \label{fig:ap_rel_change}
\end{figure}

The iterative process changes the input order of points to Slim-tree using \method's own overall ranking. The changed input order having outliers as the last insertions can improve performance as seen in Fig. \ref{fig:ap_rel_change}. The empirical results show how the AUCPR percentage gain stabilizes, approximately, in between $5\%$ and $20\%$. The stabilization happens after the first iterations. As the dashed red line denotes, from second iteration onwards, all iterations have better performance than the first one. These results confirm the efficacy of the proposed sequential refinement of rankings, and justify the choice of the default value $\maxiter = 10$ for the maximum number of iterations, thus allowing \method to work in a parameter-free manner.

\subsection{Complexity Proofs} 

\subsubsection{Time Complexity Analysis}
\label{sec:sup_time_complexity}

\begin{proof}
\textbf{Proof for Lemma~\ref{lemma:timecomplexity}:}
As shown in Algorithm~\ref{algorithm:main}, \method has two phases.
The cost of Phase~$1$ is mostly in Lines $3$ and $4$.
The former builds a Slim-tree $\tree$ for dataset $\dataset$;
the latter analyzes $\tree$ to obtain an overall ranking $\orank$.
As Slim-tree is one state-of-the-art tree-based data structure, it requires only $O(n \log n)$ time for construction.
The time to build $\orank$ is $O(n)$, since \method needs to find $\crooti$ and $\frepi$ by comparing each object $\pointi \in \dataset$ with the objects stored in only two nodes, i.e., $\rootnode$ and $\anti$, whose sizes are limited by a small constant $c$.
The sequential refinement of rankings does not affect the complexity, since $b$ is a small constant.
Thus, the total time of Phase~$1$ is $O(n \log n)$.
In Phase~$2$, the rankings $\grank$ and $\crank$ are computed together in $O(n)$ time since they both 
reuse the score $\oscorei$ from $\orank$ and calculate $\distnni$ by comparing objects from $\anti$ 
with each object $\pointi \in \dataset$ to identify $\nni$ and $\nn{\leafi}$.
Finally, local ranking $\lrank$ is computed in $O(n + |\slradiuses|^2)$ time by computing $\lradiusi$ for every object $\pointi \in \dataset$ in $O(n)$ time; then, identifying $\kneeradius$ in $O(|\slradiuses|^2)$ time.
Provided that $|\slradiuses|$ is expected to be a small constant, the total time of Phase~$2$ is $O(n)$.
It leads to the conclusion that the time complexity of \method is $O(n \log n)$.
\end{proof}

\subsubsection{Space Complexity Analysis} \label{sec:sup_space_complexity}

\begin{proof}
\textbf{Proof for Lemma~\ref{lemma:spacecomplexity}:}
The space required by \method refers to the storage of Slim-tree $\tree$ and of the four rankings $\orank$, $\grank$, $\crank$ and $\lrank$.
As Slim-tree is one state-of-the-art tree-based data structure, it requires $O(n \log n)$ space to be stored.
Each of the rankings $\orank$, $\grank$, $\crank$ and $\lrank$ are in fact a permutation of set $\{1, 2, \dots, n\}$; therefore, they require $O(n)$ space to be stored.
It leads to the conclusion that the space complexity of \method is $O(n \log n)$.
\end{proof}

\subsection{Description of Datasets}\label{sec:sup_dataset_summary}

This section details the datasets studied in our work.
Tables~\ref{tab:benchmark_datasets} describes the datasets in Testbed~$1$.
Testbeds~$2$ and $3$ are in Table~\ref{tab:base_datasets}.

\begin{table}[H]
    \centering
    \small
    \caption{\textbf{Testbed~$\mathbf{1}$:} Summary of the $20$ benchmark datasets used for overall outlier detection task.}
    \vspace{-0.1in}
    \begin{tabularx}{0.83\columnwidth}{l|c|c|c} \hline
    Dataset & Points & Outliers & Features
    \\ \hline
    Annthyroid & 7,062 & 534 & 6
    \\
    Cardio & 2,110 & 465 & 21
    \\
    Glass & 213 & 9 & 7
    \\
    Letter & 1,598 & 100 & 32
    \\
    Mammography & 7,848 & 253 & 6
    \\
    Mnist & 7,603 & 700 & 100
    \\
    Musk & 3,062 & 97 & 166
    \\
    Optdigits & 5,198 & 132 & 64
    \\
    Pageblocks & 5,393 & 510 & 10
    \\
    Pendigits & 6,870 & 156 & 16
    \\
    Satimage-2 & 5,801 & 69 & 36
    \\
    Shuttle & 1,013 & 13 & 9
    \\
    Shuttle2 & 49,097 & 3,511 & 9
    \\
    Stamps & 340 & 31 & 9
    \\
    Thyroid & 3,656 & 93 & 6
    \\
    Vowels & 1,452 & 46 & 12
    \\
    Waveform & 3,443 & 100 & 21
    \\
    WBC & 377 & 20 & 30
    \\
    WDBC & 367 & 10 & 30
    \\
    Wilt & 4,819 & 257 & 5
    \\ \hline
    \end{tabularx}
    \label{tab:benchmark_datasets}
\end{table}

\begin{table}[!t]
    \centering
    \small
    \caption{\textbf{Testbeds~$\mathbf{2}$~and~$\mathbf{3}$:} Summary of the $8$ base datasets used in for outlier annotation.}
    \vspace{-0.1in}
    \begin{tabularx}{0.8\columnwidth}{c|c|c|c} \hline
    Dataset & Points & Outliers & Features
    \\ \hline
    S1 & 24,713 & 90 & 2
    \\
    S2 & 13,832 & 120 & 5
    \\
    S3 & 10,076 & 60 & 10
    \\ \hline
    ALOI & 48,266 & 240 & 27
    \\
    Glass & 234 & 30 & 9
    \\
    skin & 14,726 & 72 & 3
    \\
    smtp & 71,566 & 357 & 3
    \\
    Waveform & 3,391 & 48 & 21
    \\ \hline
    \end{tabularx}
    \label{tab:base_datasets}
\end{table}

\subsubsection{Synthetic Data Generation for Testbed 2}\label{sec:sup_gen_synthetic}

We show in Table \ref{tab:synthetic_gen_parameters} the parameters used in our synthetic dataset generator. For every dataset variation having a specific type of outlier, the number of outliers is equal to one third of the total number of outliers.

\begin{table}[h]
    \centering
    \small
    \caption{Parameters used in the synthetic data generator; \#pts: number of points per cluster, \#clts: number of clusters, \#out: total number of outliers (each type is 1/3rd), std: standard deviation for inlier clusters, \#dim: number of features}
    \vspace{-0.1in}
    \centering
    \begin{tabularx}{0.45\textwidth}{l|c|c|c|c|c} \hline
    Dataset & \#pts & \#clts & \#out & std & \#dim
    \\ \hline
    S1\_A & 500-5000 & 10 & 90 & 10-50 & 2
    \\
    S2\_A & 500-5000 & 7 & 120 & 10-50 & 5
    \\
    S3\_A & 500-5000 & 3 & 60 & 10-50 & 10
    \\ \hline
    \end{tabularx}
    \label{tab:synthetic_gen_parameters}
\end{table}

Our synthetic data generator uses parameters like the ones shown in Table \ref{tab:synthetic_gen_parameters}. The generator creates multiple Gaussians distributed across the space. The points generated in these initial Gaussian distributions become the inlier clusters. The space in which the center of the Gaussians can fall is defined using a range bounded by the number of Gaussians times the maximum standard deviation in all Gaussians. This way, the space can accommodate all Gaussians without too much overlap between them. 

Specific types of outliers are generated like this: (1) Local outliers are generated using as Gaussians having the same mean as the inlier Gaussians, only with a magnified standard deviation; (2) Global outliers are generated using the same method as local outliers, only with a larger standard deviation; (3) For collective outliers, we base the number of collectives clusters on the desired number of collective outliers. All collective cluster have a fixed size of ten points, and randomly pick from the list of available global outliers. These selected global outliers are labeled as collective outliers. Then, a small Gaussian is generated having the global points coordinates as mean, only with a standard deviation that is a fraction of the minimum inlier Gaussian standard deviation.

\subsubsection{Realistic Data Generation for Testbed 3}\label{sec:sup_gen_realistic}

The realistic data generator starts by taking the inlier data from the dataset passed to it. The generator then fits a Gaussian Mixture Model (GMM) to the inlier points. The GMM uses a VEI setting, which means that all fitted Gaussians have a diagonal covariance matrix, varying volume and equal shape.

After the model is finished fitting to the inlier data, the generator starts simulating local and global outliers. For each Gaussian, using its mean and a five times increased standard deviation, a Gaussian cluster of points is created. The generator returns the inlier points passed to it as input and also the simulated outliers with a binary label for outlierness. 

The outlier binary label have no use in the outlier annotation task. We label local and global outlier points using their densities provided by the GMM. Lots of generated outlier points are going to be intersecting inlier clusters. Therefore, we decided to remove a sub-sample of these points, so that we eliminate outliers that have fallen under inlier clusters. The sub-sample removed corresponds to around 68\% of the data, what would be the percentage of points one standard deviation away from any Gaussian mean. 

After sub-sampling, the remaining outlier points with the lowest density become global outliers. Remaining outlier points with the highest density become local outliers. By selecting the extreme sections to create the outliers, a gap between global and local outliers is naturally formed and they become well-defined in space. Finally, collective outliers are created by taking a number of global outliers at random. The selected global outliers have their labels changed to collective. The collective clusters are generated using a small Gaussian having $0.0001$ of the minimum standard deviation across all Gaussians.

\subsection{Summary of Competitors} \label{sec:sup_competitors}

This section presents a brief summary of the algorithms as competitors when evaluating \method.
We first describe the general outlier detectors; then, we describe algorithm SRA, which is the only competitor in outlier annotation.

\subsubsection{Competitors in Outlier Detection} \label{sec:sup_competitors_outlier_detection}

The competitors in the outlier detection task are briefly described in the following: 
\cbit
\item {{\sc kNN}\xspace} algorithm ~\cite{knn} is one well-known outlier detector. Assuming that the value of a parameter $k \in \mathbb{Z}^+$ is given by the user, this algorithm ranks instances by the distances to their $k^{th}$ nearest neighbors, assuming that outliers as those objects having the farthest neighbors;
\item {{\sc LOF}\xspace} ~\cite{lof} is another well-known outlier detector. 
It receives a parameter $k \in \mathbb{Z}^+$ as input and detects the $k^{th}$ nearest neighbors of each object $\pointi \in \dataset$. The neighbors are then used to compute a local neighborhood factor that accounts for clusters with distinct densities that may exist in $\dataset$;
\item  {{\sc OCSVM}\xspace}~\cite{scholkopf2000support} aims to isolate outliers into a single cluster. This is performed by using kernels that can make the objects more linearly separable from the decision boundary that separates the inliers and the outliers;
\item  {{\sc SIF}\xspace}~\cite{czekalski2021similarity} takes the original idea of iForest~\cite{liu2008isolation} to separate the points using random cuts in space. Differently, instead of using features to perform cuts, it distances or similarities. Similarities are used to projects points onto a hypothetical line, as in Similarity Forests \cite{sathe2017similarity}. This line is defined using two random points as reference. In practice, a sub-sample of the data is divided in two. Half closer to the first point of reference, half closer to the second point of reference. This split defines two new nodes for the tree, which is going to have its nodes sequentially split into more and more nodes, just as in iForest. Many trees reproduce these same steps, building up to a forest. Each tree starts with a different sub-sample of the original dataset to avoid overfitting. Points isolated fast, in the initial nodes of the tree, have higher outlierness score.
\ceit

\subsubsection{Competitor {\sc SRA} in Outlier Annotation} \label{sec:sup_competitors_outlier_annotation}

{{\sc SRA}\xspace}\cite{nian2016auto} is the only competitor in the task of outlier annotation.
In summary, this algorithm separates the data space into two partitions and scores objects according to their similarity with the center of the partitions.
It considers only two types of outliers, namely point outliers and collective outliers.
Also, only one ranking per dataset $\dataset$ is provided, which depends on what type of outliers is predominant in $\dataset$.  
Note that point outliers may be global or local ones; distinctly from our \method, algorithm {{\sc SRA}\xspace} fails to distinguish between these two types.

\subsection{Model Configurations} \label{sec:sup_model_config}

\begin{table}
    \centering
    \small
    \caption{Hyperparameters and configurations for the baseline methods}
    \vspace{-0.1in}
    \begin{tabularx}{0.8\columnwidth}{c|c|c|c} \hline
    LOF\_k & kNN\_k & OCSVM\_nu & SIF\_n\_trees
    \\ \hline
    1 & 1 & 0.01 & 100
    \\
    3 & 3 & 0.05 & -
    \\
    5 & 5 & 0.1 & -
    \\
    10 & 10 & 0.15 & -
    \\
    15 & 15 & 0.2 & -
    \\
    30 & 30 & 0.25 & -
    \\
    50 & 50 & 0.3 & -
    \\
    75 & 75 & 0.35 & -
    \\
    100 & 100 & 0.4 & -
    \\
     & - & 0.45 & -
    \\
     & - & 0.5 & -
    \\ \hline
    \end{tabularx}
    \label{tab:sup_competitors_parameters}
\end{table}

Table~\ref{tab:sup_competitors_parameters} reports the full list of the hyperparameters used for each of our competitors. 
Column names have the form $\mathtt{method}\_\mathtt{parameter}$, where $\mathtt{method}$ is the name of the competitor, and $\mathtt{parameter}$ is the corresponding hyperparameter name. 
SIF only uses its default value of $100$ for number of trees $n\_trees$. 
OCSVM HPs range from a very small value up to its conservative default value of $0.5$ for an upperbound on the number of outliers.
Finally, both LOF and $k$NN use values of $k \in \{1, 3, 5, 10, 15, 30, 50, 75, 100 \}$ in their searches for the $k^{th}$ nearest neighbors.

\subsection{AUCPR Results on Outlier Annotation -- Synthetic and Realistic Datasets}\label{sec:sup_ap_results}


Considering the outlier annotation task, in Tables \ref{tab:ap_synthetic} and \ref{tab:ap_realistic} we present the AUCPR results for \method and for its competitor SRA. With the same performance behavior described by the corresponding AUCROC Tables~\ref{tab:auc_synthetic} and \ref{tab:auc_realistic}, we can see \method maintaining near-ideal performance. Once again, SRA performs poorly, specially in the task of annotating local outliers.


\begin{table*}[!t]
    \small
    \caption{AUCPR performance on Testbed 2 with annotated outliers. Note that
    in Sx\_A datasets with all three outlier types, type-specific rankings are evaluated against outliers of the corresponding type only.}
    \vspace{-0.1in}
    \centering
    \begin{tabularx}{0.97\textwidth}{l|c @{\cspace} c|c @{\cspace} c|c @{\cspace} c|c @{\cspace} c} \hline
    Dataset & \method-$\orank$ & SRA & \method-$\lrank$ & SRA & \method-$\grank$ & SRA & \method-$\crank$ & SRA
    \\ \hline
    S1\_A & \textbf{0.919} & 0.458 & \textbf{0.276} & 0.001 & \textbf{1} & 0.661 & \textbf{0.412} & Not flagged
    \\
    S1\_L & \textbf{0.615} & 0.001 & \textbf{0.892} & 0.001 & - & - & - & -
    \\
    S1\_G & \textbf{1} & 0.696 & - & - & \textbf{1} & 0.696 & - & -
    \\
    S1\_C & \textbf{1} & 0.667 & - & - & - & - & \textbf{1} & Not flagged
    \\
    S2\_A & \textbf{0.995} & 0.368 & \textbf{0.302} & 0.003 & \textbf{1} & 0.086 & \textbf{0.967} & Not flagged
    \\
    S2\_L & \textbf{0.945} & 0.017 & \textbf{0.996} & 0.017 & - & - & - & -
    \\
    S2\_G & \textbf{1} & 0.295 & - & - & \textbf{1} & 0.295 & - & -
    \\
    S2\_C & \textbf{1} & 0.751 & - & - & - & - & \textbf{1} & Not flagged
    \\
    S3\_A & \textbf{0.978} & 0.444 & \textbf{0.276} & 0.01 & \textbf{1} & 0.597 & \textbf{1} & Not flagged
    \\
    S3\_L & \textbf{0.816} & 0.041 & \textbf{0.922} & 0.041 & - & - & - & -
    \\
    S3\_G & \textbf{1} & 0.671 & - & - & \textbf{1} & 0.671 & - & -
    \\
    S3\_C & \textbf{1} & 0.501 & - & - & - & - & \textbf{1} & Not flagged
    \\ \hline
    \end{tabularx}
    \label{tab:ap_synthetic}
\end{table*}

\begin{table*}[!t]
    \footnotesize
    \caption{AUCPR performance on Testbed 3 with annotated outliers. Note that
    in variant \_A datasets with all three outlier types, type-specific rankings are evaluated against outliers of the corresponding type only.}
    \vspace{-0.1in}
    \begin{tabularx}{\textwidth}{l|Xl|Xl|Xl|Xl} \hline
    Dataset & \method-$\orank$ & SRA & \method-$\lrank$ & SRA & \method-$\grank$ & SRA & \method-$\crank$ & SRA
    \\ \hline
    ALOI\_A & \textbf{0.997} & 0.431 & \textbf{0.303} & Not flagged & \textbf{1} & Not flagged & \textbf{0.935} & 0.18
    \\
    ALOI\_L & \textbf{0.976} & 0.567 & \textbf{0.986} & 0.567 & - & - & - & -
    \\
    ALOI\_G & \textbf{1} & 0.399 & - & - & \textbf{1} & Not flagged & - & -
    \\
    ALOI\_C & \textbf{1} & 0.431 & - & - & - & - & \textbf{1} & 0.431
    \\
    Glass\_A & \textbf{0.87} & 0.817 & \textbf{0.144} & 0.069 & \textbf{1} & 0.418 & \textbf{1} & Not flagged
    \\
    Glass\_L & 0.241 & \textbf{0.245} & \textbf{0.396} & 0.245 & - & - & - & -
    \\
    Glass\_G & \textbf{1} & 0.882 & - & - & \textbf{1} & 0.882 & - & -
    \\
    Glass\_C & \textbf{1} & 0.521 & - & - & - & - & \textbf{1} & Not flagged
    \\
    skin\_A & \textbf{0.97} & 0.634 & \textbf{0.273} & 0.027 & \textbf{0.979} & 0.239 & \textbf{0.746} & Not flagged
    \\
    skin\_L & \textbf{0.775} & 0.071 & \textbf{0.932} & 0.071 & - & - & - & -
    \\
    skin\_G & \textbf{1} & 0.591 & - & - & \textbf{1} & 0.591 & - & -
    \\
    skin\_C & \textbf{1} & 0.987 & - & - & - & - & \textbf{1} & Not flagged
    \\
    smtp\_A & 0.909 & \textbf{0.91} & \textbf{0.229} & 0.116 & \textbf{0.991} & 0.749 & \textbf{0.391} & Not flagged
    \\
    smtp\_L & \textbf{0.608} & 0.005 & \textbf{0.775} & 0.005 & - & - & - & -
    \\
    smtp\_G & \textbf{0.996} & 0.983 & - & - & \textbf{1} & 0.983 & - & -
    \\
    smtp\_C & \textbf{0.998} & 0.257 & - & - & - & - & \textbf{1} & Not flagged
    \\
    Waveform\_A & \textbf{1} & 0.181 & \textbf{0.298} & 0.03 & \textbf{1} & 0.01 & \textbf{1} & Not flagged
    \\
    Waveform\_L & \textbf{0.888} & 0.049 & \textbf{1} & 0.049 & - & - & - & -
    \\
    Waveform\_G & \textbf{1} & 0.029 & - & - & \textbf{1} & 0.029 & - & -
    \\
    Waveform\_C & \textbf{1} & 0.362 & - & - & - & - & \textbf{1} & Not flagged
    \\ \hline
    \end{tabularx}
    \label{tab:ap_realistic}
\end{table*}

\end{document}